\documentclass[10pt,twocolumn,letterpaper]{article}

\usepackage{iccv}
\usepackage{times}
\usepackage{epsfig}
\usepackage{graphicx}
\usepackage{amsmath}
\usepackage{amssymb}

\usepackage[pagebackref=true,breaklinks=true,letterpaper=true,colorlinks,bookmarks=false]{hyperref}

\iccvfinalcopy 

\def\iccvPaperID{1854} 

\ificcvfinal\fi

\usepackage{booktabs}
\usepackage{makecell}
\usepackage{multirow}
\usepackage{bbding}
\usepackage{xcolor} 
\usepackage{pifont}
\usepackage{colortbl}
\usepackage{hyperref}
\usepackage{diagbox}
\usepackage{enumitem}
\usepackage{bm}

\newcommand{\ieno}{\textit{i}.\textit{e}.}
\newcommand{\egno}{\textit{e}.\textit{g}.}
\newcommand{\etcno}{\textit{etc.}}
\newcommand{\ours}{AFFNet}
\newcommand{\ourop}{AFF}

\newcommand{\ourblock}{AFF Block}

\begin{document}

\title{Adaptive Frequency Filters As Efficient Global Token Mixers}

\author{Zhipeng Huang\textsuperscript{\rm 1,2}\thanks{This work was done when Zhipeng Huang was an intern at Microsoft Research Asia.} \quad 
Zhizheng Zhang\textsuperscript{\rm 2}\quad 
Cuiling Lan\textsuperscript{\rm 2}\quad 
Zheng-Jun Zha\textsuperscript{\rm 1}\quad 
Yan Lu\textsuperscript{\rm 2}\quad 
Baining Guo\textsuperscript{\rm 2}\\
\textsuperscript{\rm 1}{University of Science and Technology of China}\quad
\textsuperscript{\rm 2}{Microsoft Research Asia}\\
\tt\small \{zhizzhang,\ culan,\ yanlu,\ bainguo\}@microsoft.com \\ 
\tt\small hzp1104@mail.ustc.edu.cn \quad
\tt\small zhazj@ustc.edu.cn}


\maketitle
\ificcvfinal\thispagestyle{empty}\fi

\begin{abstract}

Recent vision transformers, large-kernel CNNs and MLPs have attained remarkable successes in broad vision tasks thanks to their effective information fusion in the global scope. However, their efficient deployments, especially on mobile devices, still suffer from noteworthy challenges due to the heavy computational costs of self-attention mechanisms, large kernels, or fully connected layers. In this work, we apply conventional convolution theorem to deep learning for addressing this and reveal that adaptive frequency filters can serve as efficient global token mixers. With this insight, we propose Adaptive Frequency Filtering (AFF) token mixer. This neural operator transfers a latent representation to the frequency domain via a Fourier transform and performs semantic-adaptive frequency filtering via an elementwise multiplication, which mathematically equals to a token mixing operation in the original latent space with a dynamic convolution kernel as large as the spatial resolution of this latent representation. We take AFF token mixers as primary neural operators to build a lightweight neural network, dubbed AFFNet. Extensive experiments demonstrate the effectiveness of our proposed AFF token mixer and show that AFFNet achieve superior accuracy and efficiency trade-offs compared to other lightweight network designs on broad visual tasks, including visual recognition and dense prediction tasks.

\end{abstract}

\section{Introduction}

Remarkable progress has been made in ever-changing vision network designs to date, wherein effective token mixing in the global scope is constantly highlighted. Three existing dominant network families, \ieno, Transformers, CNNs and MLPs achieve global token mixing with their respective ways. Transformers \cite{dosovitskiy2020image,touvron2021training,liu2021swin,chu2021twins,yu2022metaformer} mix tokens with self-attention mechanisms where pairwise correlations between query-key pairs are taken as mixing weights. CNNs achieve competitive performance with transformers by scaling up their kernel sizes \cite{peng2017large,ding2022scaling,liu2022more,chen2022scaling}. MLPs \cite{tolstikhin2021mlp,hou2022vision,lian2021mlp} provide another powerful paradigm via fully connections across all tokens. All of them are effective but computationally expensive, imposing remarkable challenges in practical deployments, especially on edge devices.

\begin{figure}[t]
    \centering
    \includegraphics[width=0.46\textwidth]{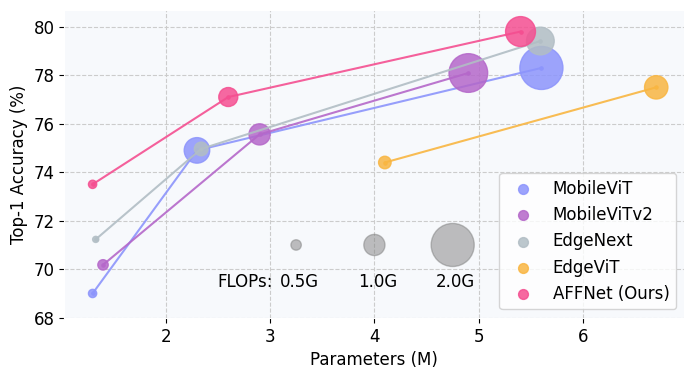}
    \caption{Comparison of Top-1 accuracy on ImageNet-1K~\cite{russakovsky2015imagenet} between our proposed \ours~to some state-of-the-art lightweight networks that have global token mixing. The bubble size corresponds to FLOPs.}
    \label{fig:paramflops-acc}
\end{figure} 

Recently, there is increased attention on improving the efficiency of token mixing in transformers. Some works ~\cite{kitaev2020reformer,liu2021swin,chu2021twins,hassani2022dilated,mehta2021mobilevit,pan2022fast,li2022next} squeeze the scope of token mixing in different ways to compromise the representation capacities of neural networks for their efficiencies. Other works reduce the complexity of the matrix operations in self-attention by making use of the associativity property of matrix products \cite{katharopoulos2020transformers} or low-rank approximation methods \cite{guo2019low,xiong2021nystromformer}. These methods all sacrifice the expressiveness of neural networks and lead to unsatisfactory performance of efficient network designs. A general-purpose global token mixing for lightweight networks is still less explored. Better trade-off between accuracy and efficiency for global-scope token mixing is worthy of further study. 

In this work, we reveal that \textit{adaptive frequency filters can serve as efficient global token mixers}, inspired by the convolution theorem \cite{mcgillem1991continuous,rabiner1975theory,oppenheim1999discrete} widely used in conventional signal processing. 
This theorem states that a convolution in one domain mathematically equals the Hadamard product (also known as elementwise product) in its corresponding Fourier domain. This equivalence allows us to frame global token mixing as a large-kernel convolution in the latent space and efficiently implement this convolution with a Hadamard product operation in the frequency domain by performing Fourier transforms on tokens in the latent space.

Besides large scopes, the adaptability to semantics also matters for token mixing as studied in \cite{dai2017deformable,chen2020dynamic,activemlp,bau2020understanding,Wu_2021_CVPR}. This means that the weights for token mixing should be instance-adaptive. Moreover, different semantic attributes of the learned latent representations distribute in different channels \cite{bau2020understanding,wu2021stylespace}. This property poses requirements for channel-specific token mixing wherein the weights of token mixing vary across different channels. From the perspective of framing global adaptive token mixing as a convolution, the kernel of this convolution operation should be not only large but also spatially dynamic. However, it is well known that dynamic convolutions are computationally expensive in common. Large-kernel dynamic convolutions seem extremely prohibitive for efficient/lightweight network designs. In this paper, we propose to adopt frequency filtering in the Fourier domain with learned instance-adaptive masks as a mathematical equivalent of token mixing using large-kernel dynamic convolutions by making use of the aforementioned convolution theorem. This equivalent could reduce the complexity of token mixing from $\mathcal{O}(N^2)$ to $\mathcal{O}(N \log N)$ thanks to adopting Fast Fourier Transforms (FFT), which is more computationally efficient.

With the key insight above, we propose Adaptive Frequency Filtering (\ourop) token mixer. In this neural operator, the latent representations (\ieno, a set of tokens) are transferred from its original latent space to a frequency space via a 2D discrete Fourier transform applied spatially. In this way, we get the frequency representations whose spatial positions correspond to different frequency components. We adopt an extremely lightweight network to learn instance-adaptive masks from these frequency representations, and then calculate the Hadamard product between the learned masks and the frequency representations for adaptive frequency filtering. The filtered representations are transferred back to the original latent space via an inverse Fourier transform. The features after this inverse transform could be viewed as the results of token mixing with depthwise convolution kernels whose spatial dimensions are as large as those of latent representations (\ieno, the token set). According to the convolution theorem \cite{mcgillem1991continuous}, our proposed operation mathematically equals to taking the tensors of applying an inverse Fourier transform to the learned masks in the Fourier domain as the corresponding kernel weights and perform convolution with this kernel in the original domain. Detailed introduction, demonstration and analysis are given in subsequent sections.

Furthermore, we take the proposed \ourop~token mixer as the primary neural operator and assemble it into an \ourop~block together with a plain channel mixer. \ourop~blocks serve as the basic units for constructing efficient vision backbone, dubbed \ours. We evaluate the effectiveness and efficiency of our proposed \ourop~token mixer by conducting extensive ablation study and comparison across diverse vision tasks and model scales.

Our contributions can be summarized in the following:

\begin{itemize}[noitemsep,nolistsep,leftmargin=*]
\item We reveal that adaptive frequency filtering in the latent space can serve as efficient global token mixing with large dynamic kernels, and propose Adaptive Frequency Filtering (\ourop) token mixer.
\item We conduct theoretical analysis and empirical study to compare our proposed \ourop~token mixer with other related frequency-domain neural operators from the perspective of information fusion for figuring out what really matters for the effects of token mixing.
\item We take \ourop~token mixer as the primary neural operator to build a lightweight vision backbone \ours. \ours~achieves the state-of-the-art accuracy and efficiency trade-offs compared to other lightweight network designs across a broad range of vision tasks. An experimental evidence is provided in Fig.\ref{fig:paramflops-acc}.
\end{itemize}

\section{Related Work}

\subsection{Token Mixing in Deep Learning}

Mainstream neural network families, \ieno, CNNs, Transformers, MLPs, differ in their ways of token mixing, as detailed in \cite{wei2022activemlp}. CNNs \cite{o2015introduction} mix tokens with the learned weights of convolution kernels where the spatial kernel size determines the mixing scope. Commonly, the weights are deterministic and the scope is commonly a local one. Transformers \cite{vaswani2017attention,dosovitskiy2020image} mix tokens with pairwise correlations between query and key tokens in a local \cite{liu2021swin,chu2021twins} or global\cite{dosovitskiy2020image,touvron2021training} range. These weights are semantic-adaptive but computationally expensive due to the $\mathcal{O}(N^2)$ complexity. MLPs commonly mix tokens with deterministic weights in manually designed scopes \cite{chen2022cyclemlp,touvron2021resmlp,tolstikhin2021mlpmixer,zhang2021morphmlp} wherein the weights are the network parameters. This work aims to design a generally applicable token mixer for lightweight neural networks with three merits: computation-efficient, semantic-adaptive and effective in the global scope.

\subsection{Lightweight Neural Networks}
Lightweight neural network designs have been of high values for practical deployments.
CNNs, Transformers, and MLPs have their own efficient designs. MobileNets series \cite{howard2017mobilenets,sandler2018mobilenetv2,howard2019searching} introduce depthwise and pointwise convolutions as well as modified architectures for improving the efficiency. Shufflenet series \cite{zhang2018shufflenet,ma2018shufflenet} further improve pointwise convolution via shuffle operations.  MobileViT \cite{mehta2021mobilevit} combines lightweight MobileNet block and multi-head self-attention blocks. Its follow-up versions further improve it with a linear-complexity self-attention method \cite{mobilevitv2}. Besides, there are many works reducing the complexity of self-attention via reducing the region of token mixing \cite{liu2021swin,chu2021twins,pan2022fast,li2022next} or various mathematical approximations \cite{guo2019low,xiong2021nystromformer,ma2021luna}. Many efficient MLPs limit the scope of token mixing to horizontal and vertical stripes \cite{zhang2022morphmlp,hou2022vision,tang2022sparse} or a manually designed region \cite{chen2021cyclemlp}.

\subsection{Frequency-domain Deep Learning}
\label{sec: related-freq}
Frequency-domain analysis has been a classical tool for conventional signal processing \cite{baxes1994digital,pitas2000digital} for a long time. Recently, frequency-domain methods begin to be introduced to the field of deep learning for analyzing the optimization \cite{xu2019training,yin2019fourier} and generalization \cite{wang2020high,xu2018understanding} capabilities of Deep Neural Networks (DNNs). Besides these, frequency-domain methods are integrated into DNNs to learn non-local \cite{chi2020fast,rao2021global,li2020fourier,guibas2021adaptive} or domain-generalizable \cite{lin2022deep} representations. Our proposed method might be similar to them at first glance but actually differs from them in both modelling perspectives and architecture designs. These five works propose different frequency-domain operations by introducing convolutions \cite{chi2020fast}, elementwise multiplication with trainable network parameters \cite{rao2021global}, matrix multiplication with trainable parameters \cite{li2020fourier}, groupwise MLP layers \cite{guibas2021adaptive} and elementwise multiplication with spatial instance-adaptive masks \cite{lin2022deep} to frequency-domain representations, respectively. All of them are not designed for the same purpose with ours. We provide detailed mathematical analysis on their shortcomings as token mixers and conduct extensive experimental comparisons in the following sections.

\begin{figure*}[!t]
	\centering
	\includegraphics[width=0.96\textwidth]{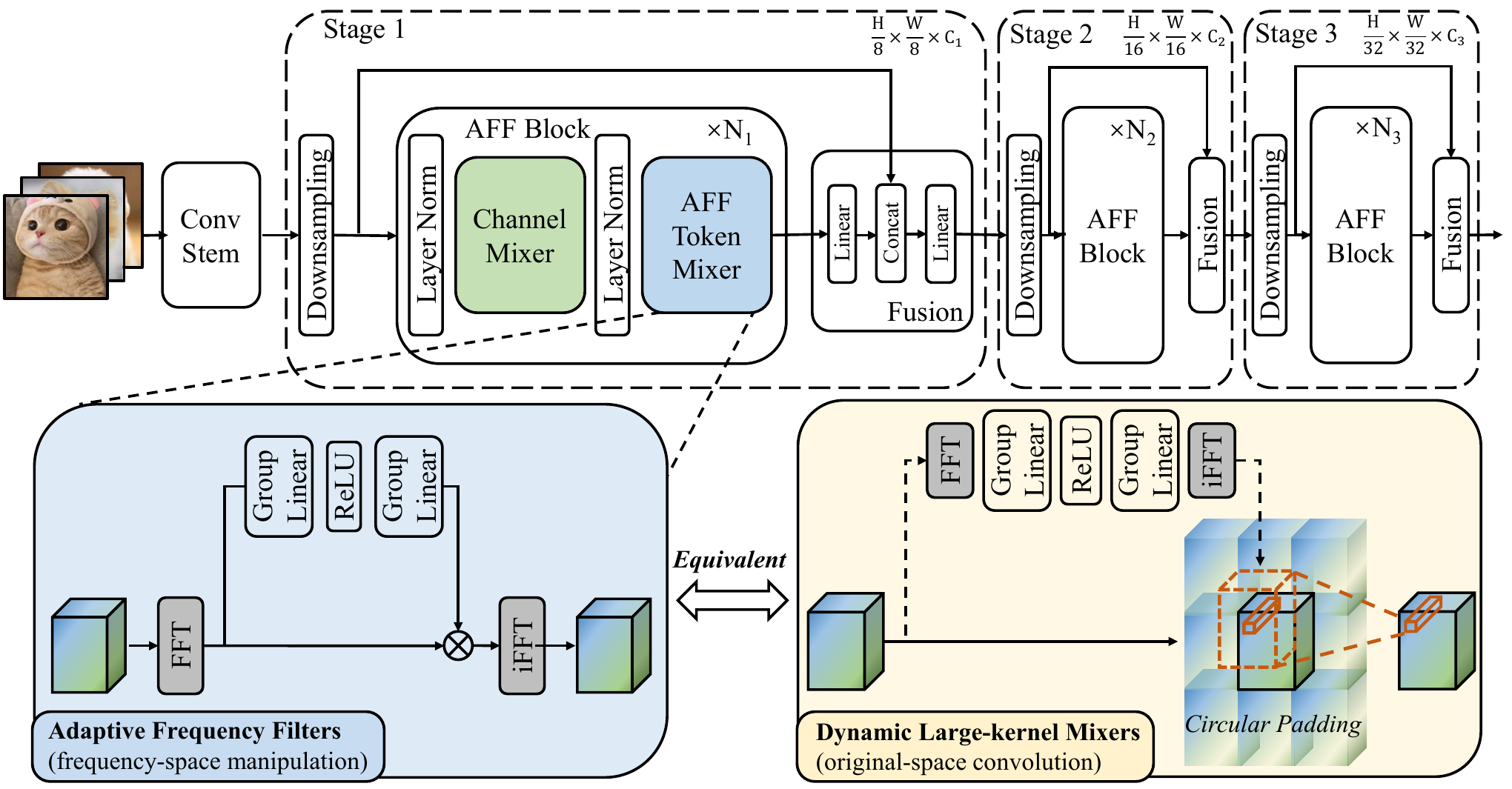}
	\caption{Illustration of our proposed \ourop~token mixer and its corresponding network \ours. The \ourop~token mixer is implemented by adaptive frequency filters at the bottom left and mathematically equals to the mixing operation at the bottom right. This operation can be viewed as token mixing with a large-kernel dynamic convolution where the kernel weights are inferred by the sub-network as shown in the bottom right sub-figure.}
	\label{fig:framework}
\end{figure*}

\section{Method}

We first describe a unified formulation of token mixing, then introduce our proposed Adaptive Frequency Filtering (\ourop) token mixer. We further analyze what properties matter for a frequency-domain operation in terms of its effects on token mixing. We finally introduce \ours~which is a lightweight backbone with \ourop~token mixer as its core.

\subsection{Unified Formulation of Token Mixing}

Token mixing is of high importance since learning non-local representations is critical for visual understanding \cite{wang2018non,dosovitskiy2020image,touvron2021training}. In most mainstream neural networks, the input image is firstly patchified into a feature tensor $\mathbf{X}\in \mathbb{R}^{H \times W \times C}$ whose spatial resolution is $H\times W$ and the number of channels is $C$. This feature tensor could be viewed as a set of tokens, in which each token can be denoted as $\mathbf{x}\in \mathbb{R}^{1 \times 1 \times C}$. The updated token for a query $\mathbf{x}^q$ after token mixing in its contextual region $\mathcal{N}(\mathbf{x}^q)$ can be formulated in a unified form:
\begin{equation}
    \hat{\mathbf{x}^q} = \sum_{i\in\mathcal{N}(\mathbf{x}^q)} 
    \bm{\omega}^{i\rightarrow q} \times \phi(\mathbf{x}^i), 
    \label{eq: token-mix}
\end{equation}
where $\hat{\mathbf{x}^q}$ refers to the updated $\mathbf{x}^q$ and $\mathbf{x}^i$ refers to the tokens in $\mathcal{N}(\mathbf{x}^q)$. $\phi(\cdot)$ denotes the embeding functions. $\bm{\omega}^{i\rightarrow q}$ represents the weights of information fusion from token $\mathbf{x}^i$ to the updated $\mathbf{x}^q$. The symbol $\times$ could be Hadamard product or matrix multiplication.

We revisit the prevailing token mixing methods in different types of network architectures in terms of their effectiveness and efficiency. For CNNs, tokens are mixed by matrix multiplication with deterministic network parameters as the mixing weights. Here, the kernel sizes of convolutions determine the scopes of token mixing. This makes mixing in a global scope quite costly due to the quadratically increased parameters and FLOPs as the kernel size increases. Transformers mix tokens with pairwise correlations between query and key tokens. Its computational complexity is $\mathcal{O}(N^2)$ ($N$ is the total number of tokens), limiting its applications in lightweight networks. Like CNNs, MLPs also mix tokens with deterministic network parameters. The scope of token mixing in advanced MLPs \cite{chen2022cyclemlp,touvron2021resmlp,tolstikhin2021mlpmixer,zhang2021morphmlp} are commonly manually design, where the globality comes at the cost of huge computational complexity. They are all not specifically designed for lightweight neural networks.

This work aims to deign a \textit{computationally efficient}, \textit{semantically adaptive} and \textit{global-scope} token mixer for lightweight networks. This requires a large $\mathcal{N}(\mathbf{x}^q)$ and instance-adaptive $\bm{\omega}^{i\rightarrow q}$ with less network parameters and low computation costs as possible.

\subsection{Adaptive Frequency Filtering Token Mixer}
\label{sec:AFF-token-mixer}

We apply the convolution theorem \cite{mcgillem1991continuous,rabiner1975theory,oppenheim1999discrete} to deep learning for designing a token mixer with aforementioned merits for lightweight neural networks. 
Based on this theorem, we reveal that \textit{adaptive frequency filters can serve as efficient global token mixers}. In the following, we introduce its mathematical modelling, architecture design and the equivalence between them for our proposed token mixer.

\paragraph{Modelling.} To simplify understanding, we frame token mixing in the form of global convolution, succinctly denoted by $\mathbf{\hat{X}}=\mathcal{K}*\mathbf{X}$. For the query token at position $(h, w)$, \ieno, $\mathbf{X}(h, w)$, Eq.(\ref{eq: token-mix}) can be reformulated as:
\begin{equation}
    \mathbf{\hat{X}}(h, w) =\!\sum_{h'=-\lfloor \frac{H}{2}  \rfloor}^{\lfloor \frac{H}{2} \rfloor} \sum_{w'=-\lfloor \frac{W}{2} \rfloor}^{\lfloor \frac{W}{2} \rfloor} \mathcal{K}(h', w') \mathbf{X}(h-h', w-w'),
    \label{eq: modelling-1}
\end{equation}
where $\mathbf{\hat{X}}(h, w)$ represents the updated token for $\mathbf{X}(h, w)$ after token mixing. $H$ and $W$ are the height and weight of the input tensor, respectively. $\mathcal{K}(h', w')$ denotes the weights for token mixing, implemented by a global convolution kernel which has the same spatial size with $\mathbf{X}$. The padding operation for $\mathbf{X}$ is omitted here for simplicity and the specific padding method is introduced in the subsequent parts.

With the expectation for our proposed token mixer as a \textit{semantic-adaptive} and \textit{global-scope} one, the weights $\mathcal{K}$ for token mixing should be adaptive to $\mathbf{X}$ and of large spatial size. As illustrated by the lower right subfigure in Fig.\ref{fig:framework}, a straightforward way for enabling $\mathcal{K}$ adaptive to $\mathbf{X}$ is to implement it with a dynamic convolution kernel \cite{jia2016dynamic,chen2020dynamic,he2019dynamic,zhang2021accurate}, \ieno, inferring weights of $\mathcal{K}$ with $\mathbf{X}$ as the inputs of a sub-network. However, adopting dynamic convolutions is usually computational costly, even more so, when using large-kernel ones. This thus imposes big challenges in designing an efficient token mixer for lightweight networks along this way. Next, we introduce an efficient method as its equivalent implementation by making use of the convolution theorem \cite{mcgillem1991continuous}.

\paragraph{Architecture.} The convolution theorem \cite{mcgillem1991continuous,oppenheim1999discrete,rabiner1975theory} for inverse Fourier transform  states that a convolution in one domain mathematically equals the Hadamard product in its corresponding Fourier domain. This inspires us to propose a lightweight and fast architecture (illustrated by the lower left part of Fig.\ref{fig:framework}) as an extremely efficient implementation of our modelling above.

Given feature $\mathbf{X} \in \mathbb{R}^{H\times W \times C}$, \ieno, a set of tokens in the latent space, we adopt Fast Fourier Transform (FFT) to obtain the corresponding frequency representations $\mathbf{X}_{F}$ by $\mathbf{X}_{F} = \mathcal{F}(\mathbf{X})$. The detailed formulation of $\mathcal{F}(\cdot)$ is:
\begin{equation}
    \mathbf{X}_{F} (u, v) = \sum_{h=0}^{H-1} \sum_{w=0}^{W-1} \mathbf{X} (h, w) e^{-2\pi i(uh+vw)}.
    \label{eq: fft}
\end{equation}
As indicated by Eq.(\ref{eq: fft}), features of different spatial positions in $\mathbf{X}_{F}$ correspond to different frequency components of $\mathbf{X}$. They incorporate global information from $\mathbf{X}$ with a transform of $\mathcal{O}(N\log N)$ complexity.


We apply the aforementioned convolution theorem to achieve efficient global token mixing for $\mathbf{X}$ by filtering its frequency representation $\mathbf{X}_{F}$ with a learnable instance-adaptive mask. We further adopt inverse FFT to the filtered $\mathbf{X}_{F}$ for getting the updated feature representations $\mathbf{\hat{X}}$ in the original latent space. This process can be formulated as:
\begin{equation}
    \mathbf{\hat{X}} = \mathcal{F} ^{-1} [\mathcal{M}(\mathcal{F} (\mathbf{X})) \odot \mathcal{F} (\mathbf{X})],
    \label{eq: aff}
\end{equation}
where $\mathcal{M}(\mathcal{F}(\mathbf{X}))$ is the mask tensor learned from $\mathbf{X}_{F}$, which has the same shape with $\mathbf{X}_{F}$. As shown in the lower left subfigure in Fig.\ref{fig:framework}, to make the network lightweight as possible, $\mathcal{M}(\cdot)$ is efficiently implemented by a group $1\times 1$ convolution (linear) layer, followed by a ReLU function and another group linear layer. $\odot$ denotes Hadamard product, also known as elementwise multiplication, and $\mathcal{F}^{-1}(\cdot)$ denotes inverse Fourier transform. Here, $\mathbf{\hat{X}}$ can be viewed as the results of global adaptive token mixing for $\mathbf{X}$, which is mathematically equivalent to adopting a large-size dynamic convolution kernel as the weights for token mixing. The equivalence is introduced in the following.

\paragraph{Equivalence.}
The convolution theorem still applies to the latent representations of neural networks. The multiplication of two signals in the Fourier domain equals to the Fourier transform of a convolution of these two signals in their original domain. When applying this to the frequency-domain multiplication in Fig.(\ref{fig:framework}), we know that:
\begin{equation}
    \mathcal{M}(\mathcal{F}(\mathbf{X})) \odot \mathcal{F} (\mathbf{X}) = \mathcal{F}\{\mathcal{F} ^{-1}[\mathcal{M}(\mathcal{F} (\mathbf{X}))] \ast \mathbf{X}\}.
    \label{eq: conv-theorem}
\end{equation}
Combining Eq.(\ref{eq: aff}) and Eq.(\ref{eq: conv-theorem}), it is easy to get that:
\begin{equation}
    \mathbf{\hat{X}} = \mathcal{F}^{-1}[\mathcal{M}(\mathcal{F} (\mathbf{X}))] \ast \mathbf{X},
    \label{eq: equivalent-token-mix}
\end{equation}
where $\mathcal{F}^{-1}(\mathcal{M}(\mathcal{F} (\mathbf{X})))$ is a tensor of the same shape with $\mathbf{X}$, which could be viewed as a dynamic depthwise convolution kernel as large as $\mathbf{X}$ in spatial. This kernel is adaptive to the contents of $\mathbf{X}$. Due to the property of Fourier transform \cite{mcgillem1991continuous}, a circular padding is adopted to $\mathbf{X}$ here as shown in Fig.\ref{fig:framework}. So far, we understand why the operation in Eq.(\ref{eq: aff}) mathematically equals to a global-scope token mixing operation with semantic-adaptive weights.

\subsection{Analysis}
\label{sec:analysis}

As introduced in Sec.\ref{sec: related-freq}, there have been some studies applying frequency-domain methods to DNN for learning non-local or domain-generalizable representations in previous works \cite{chi2020fast,rao2021global,li2020fourier,guibas2021adaptive,lin2022deep}. They are all designed for different purposes with ours. In this section, we revisit the frequency-domain operations in these works from the perspective of token mixing and compare our design with them.


FFC~\cite{chi2020fast} and AFNO~\cite{guibas2021adaptive} adopt linear (also known as $1\!\times\!1$ convolution) layers with non-linear activation functions to the representations in the frequency domain. Specifically, AFNO~\cite{guibas2021adaptive} adopt a linear layer followed by a ReLU function, another linear layer and a SoftShrink\footnote{\url{https://pytorch.org/docs/stable/generated/torch.nn.Softshrink.html}} function to the frequency representations after Fourier transforms, which can be briefly described as \textit{FFT$\to$Linear$\to$ReLU$\to$Linear$\to$SoftShrink$\to$iFFT}. Here, linear layer and Fourier transform are in fact commutative, \ieno, $\operatorname{Linear}(\mathcal{F}(\mathbf{X}))\!=\!\mathcal{F}(\operatorname{Linear}(\mathbf{X}))$, which can be proved with the distributive property of matrix multiplication by:
\begin{equation}
\begin{split}
    &\ \mathbf{W}_{Linear} \sum_{h=0}^{H-1} \sum_{w=0}^{W-1} \mathbf{X}(h, w) e^{-2\pi i(uh+vw)} \\
    = &\ \sum_{h=0}^{H-1} \sum_{w=0}^{W-1} \left(\mathbf{W}_{Linear} \mathbf{X}(h, w)\right) e^{-2\pi i(uh+vw)},
\end{split}
\label{eq: commutative}
\end{equation}
where $\mathbf{W}_{Linear}$ denotes the parameters of a linear layer. We know that successive Fourier transform and its inverse transform equal to an identity function. Thus, the architecture of AFNO could be rewrote as: \textit{FFT$\to$Linear$\to$ReLU$\to$\\(iFFT$\to$FFT)$\to$Linear$\to$SoftShrink$\to$iFFT}. Upon the commutative law proved in Eq.(\ref{eq: commutative}), we can know this architecture is in fact equivalent to \textit{Linear$\to$\textbf{FFT$\to$ReLU$\to$iFFT}\\$\to$Linear$\to$\textbf{FFT$\to$SoftShrink$\to$iFFT}}. Now, it is easy to find that only ReLU and SoftShrink functions remain in the Fourier domain. These two deterministic functions cannot achieve semantic-adaptive filtering as our proposed \ourop~token mixer does. The same problem also exists in FFC~\cite{chi2020fast}.

GFNet~\cite{rao2021global} and FNO~\cite{li2020fourier} multiply the representations after Fourier transforms with trainable network parameters. GFNet~\cite{rao2021global} adopts elementwise multiplication while FNO~\cite{li2020fourier} uses matrix multiplication. Both of them are not semantic-adaptive since the masks implemented by network parameters are shared over different instances and fixed after training. Besides, they cannot support for variable-size inputs since the shapes of these masks are fixed, leading to the lack of flexibility in their practical using.

\begin{table}[t]
  \centering
  \resizebox{\linewidth}{!}{
    \begin{tabular}{l|cccccc}
    \toprule
    \multicolumn{1}{c|}{Properties} & FFC & AFNO & \!GFNet\! & FNO & DFF & Ours \\
    \midrule
    Semantic-adaptive &   \ding{55}     & \ding{55}   &    \ding{55}   &   \ding{55}    &   \ding{51}    & \ding{51} \\
    Variable-size input & \ding{51} & \ding{51} & \ding{55}  & \ding{55}  & \ding{51} & \ding{51} \\
    Channel-wise mix &  \ding{51}      &  \ding{51}     &   \ding{51}    &   \ding{51}    &  \ding{55}     & \ding{51} \\
    \bottomrule
    \end{tabular}%
    }
    \vspace{+1.6mm}
    \caption{Comparisons of our proposed \ourop~token mixer with other frequency-domain neural operators in terms of three important properties for token mixing.}
  \label{tab:compare-operations}%
\end{table}%

DFF~\cite{lin2022deep} learns a spatial mask to filter out frequency components that are not conductive to domain generalization. It is proposed for domain generalization problems in which only spatial mask is needed as studied in~\cite{lin2022deep} since different spatial position of the features after a Fourier transform correspond to different frequency components. However, it is not competent as a token mixer since the learned mask is shared along the channel dimension. This means that the weights for its equivalent token mixing are shared for different channels. However, different channels commonly represent different semantic attributes \cite{bau2020understanding,wu2021stylespace}, thus requiring adaptive weights in token mixing.

We summarize the comparisons of different frequency-domain designs in terms of three important properties for token mixing in Table \ref{tab:compare-operations}. The results of experimental verification are in Table \ref{tab:freq_operations} as follows.

\subsection{Network Architectures}

With our \ourop~token mixer as the core neural operator, we introduce its corresponding module and network design.

\paragraph{\ourblock}
For the output $\mathbf{X}^{l-1}$ of the ($l\!-\!1$)-th \ourblock, we adopt the commonly used module MBConv~\cite{mobilevitv2,tan2019efficientnet,tan2021efficientnetv2,mehta2021mobilevit,tu2022maxvit} with Layer Normalization (LN) for channel mixing, then feed it to our proposed \ourop~token mixer for global token mixing to get the output of $l$-th \ourop~block. Skip-connections for channel mixing and token mixing are adopted to facilitate model training. The entire architecture of \ourblock~can be formulated as:

\begin{equation}
\begin{split}
    & \hat{\mathbf{X}}^l=\operatorname{MBConv}^l\left(\operatorname{LN}\left(\mathbf{X}^{l-l}\right)\right)+\mathbf{X}^{l-l} \\
    & \mathbf{X}^l=\operatorname{\ourop}^l\left(\operatorname{LN}\left(\hat{\mathbf{X}}^l\right)\right)+\hat{\mathbf{X}}^l
\end{split}
\label{eq: block}
\end{equation}


\paragraph{\ours} 
We stack multiple \ourop~blocks for constructing a lightweight backbone network, namely \ours, as shown in Fig.\ref{fig:framework}. Following the common practices \cite{mehta2021mobilevit,mobilevitv2}, we employ a convolution stem for tokenization and a plain fusion for combining local and global features at each stage. We build three versions of \ours~with different numbers of channels, yielding different parameter scales. \ours~and its tiny (\ours-T) and extremely tiny (\ours-ET) versions have 5.5M, 2.6M and 1.4M parameters, respectively. Their detailed configurations are in the Supplementary.


\section{Experiments}

We evaluate our proposed \ourop~token mixer by conducting comparisons with the state-of-the-art lightweight networks and extensive ablation studies for its design.


\subsection{Image Classification}

\paragraph{Settings.} We train different versions of our proposed lightweight networks \ours~as backbones on ImageNet-1k dataset~\cite{russakovsky2015imagenet} from scratch. All models are trained for 300 epochs on 8 NVIDIA V100 GPUs with a batch size of 1024. More implementation details are in the Supplementary.


\begin{table}[h]
  \centering
    \resizebox{\linewidth}{!}{
    \begin{tabular}{l|c|c|cc|c}
    \toprule
    \multicolumn{1}{c|}{Model} & Pub.  & Res.  & \makecell[c]{Param.  \\ (M)} & \makecell[c]{FLOPs \\ (G)} & Top-1 \\
    \midrule
    MNetv1-0.5~\cite{howard2017mobilenets} & arXiv17 & $224^2$ & 1.3 & 0.2 & 63.7 \\
    MViT-XXS~\cite{mehta2021mobilevit} & ICLR22 & $256^2$ & 1.3   & 0.4   & 69.0 \\
    EdgeNext-XXS~\cite{maaz2023edgenext} & ECCV22 & $256^2$ & 1.3 & 0.3 & 71.2 \\
    MViTv2-0.5~\cite{mobilevitv2} & TMLR23 & $256^2$ & 1.4   & 0.5   & 70.2 \\
    \rowcolor[gray]{0.9}\ours-ET  & -     & $256^2$ & 1.4   & 0.4   & 73.0 \\
    \midrule
    MNetv3-L-0.5~\cite{howard2019searching} & ICCV19 & $224^2$ & 2.6 & 0.1 & 68.8 \\
    MFormer-52~\cite{chen2022mobile}  & CVPR22 & $224^2$ & 3.6   & 0.1   & 68.7 \\
    PVTv2-B0~\cite{wang2022pvt} & CVM22 & $224^2$ & 3.7 & 0.6 & 70.5 \\
    MViT-XS~\cite{mehta2021mobilevit} & ICLR22 & $256^2$ & 2.3   & 1.0   & 74.8 \\
    EdgeNext-XS~\cite{maaz2023edgenext} & ECCV22 & $256^2$ & 2.3 & 0.5 & 75.0 \\
    EFormer-S0~\cite{li2022rethinking} & arXiv22 & $224^2$ & 3.5 & 0.4 & 75.7  \\
    MViTv2-0.75~\cite{mobilevitv2} & TMLR23 & $256^2$ & 2.9   & 1.0   & 75.6 \\
    \rowcolor[gray]{0.9}\ours-T  & -     & $256^2$ & 2.6   & 0.8   & 77.0 \\
    \midrule
    MNetv2~\cite{sandler2018mobilenetv2} & CVPR18 & $224^2$ & 6.9   & 0.6   & 74.7 \\
    ShuffleNetV2~\cite{ma2018shufflenet} & ECCV18 & $224^2$ & 5.5   & 0.6   & 74.5 \\
    MNetv3~\cite{howard2019searching} & ICCV19 & $224^2$ & 5.4   & 0.2   & 75.2 \\
    T2T-ViT~\cite{yuan2021tokens} & ICCV21 & $224^2$ & 6.9   & 1.8   & 76.5 \\
    DeiT-T~\cite{touvron2021training} & ICML21 & $224^2$ & 5.7   & 1.3   & 72.2 \\
    CoaT-Lite-T~\cite{dai2021coatnet} & ICCV21 & $224^2$ & 5.7   & 1.6   & 77.5 \\
    LeViT-128~\cite{graham2021levit} & ICCV21 & $224^2$ & 9.2   & 0.4   & 78.6 \\
    GFNet-Ti~\cite{rao2021global} & NeurIPS21 & $224^2$ & 7.0   & 1.3   & 74.6 \\
    EFormer-L1~\cite{li2022efficientformer} & NeurIPS22    & $224^2$ & 12.3  & 1.3   & 79.2 \\
    EFormer-S1~\cite{li2022rethinking} & arXiv22 & $224^2$ & 6.1 & 0.7 & 79.0 \\
    Mformer~\cite{chen2022mobile} & CVPR22 & $224^2$ & 9.4   & 0.2   & 76.7 \\
    EfficientViT~\cite{cai2022efficientvit}  & arXiv22 & $224^2$ & 7.9   & 0.4   & 78.6 \\
    EdgeViT-XS~\cite{chen2022edgevit} & ECCV22 & $256^2$ & 6.7   & 1.1   & 77.5 \\
    MOne-S3~\cite{vasu2022improved} & arXiv22 & $224^2$ & 10.1  & 1.9   & 78.1 \\
    MViT-S~\cite{mehta2021mobilevit} & ICLR22 & $256^2$ & 5.6   & 2.0   & 78.4 \\
    EdgeNext-S~\cite{maaz2023edgenext} & ECCV22 & $256^2$ & 5.6 & 1.3 & 79.4 \\
    MViTv2-1.0~\cite{mobilevitv2} & TMLR23 & $256^2$ & 4.9   & 1.8   & 78.1 \\
    \rowcolor[gray]{0.9}\ours  &   -    & $256^2$ & 5.5   & 1.5   & 79.8 \\
    \bottomrule
    \end{tabular}}%
    \vspace{+1.6mm}
  \caption{Comparisons of our proposed \ours~with other state-of-the-art lightweight networks on ImageNet-1K classification over different model scales (\ieno, $<$2M, 2M $\sim$ 4M and $>$ 4M). For conciseness, Pub., Res., Param., MNet, MOne, MFormer, EFormer and MViT are short for Publication, Resolution, Parameters, MobileNet, MobileOne, MobileFormer, EfficientFormer and MobileViT, respectively.}
  \label{tab:sota_imagenet}%
\end{table}%

\vspace{-2mm}
\paragraph{Results.}  We report the comparison results between our proposed \ours~and other SOTA lightweight models in Table \ref{tab:sota_imagenet}. We observe that our \ours~outperforms other lightweight networks with comparable model sizes in Top-1 accuracy. The \ours~reaches 79.8\% Top-1 accuracy with 5.5M parameters and 1.5G FLOPs. Our extremely tiny model \ours-ET attains 73\% Top-1 accuracy with sorely 1.4M and 0.4G FLOPs. As a result, \ours~achieves the best trade-offs between accuracy and efficiency. To show the comparison results more intuitively, we illustrate the accuracy and efficiency trade-offs of our \ours~and some advanced lightweight models with global token mixers in Fig.~\ref{fig:paramflops-acc}. Thanks to \ourop~token mixer, \ours~is superior to them by a clear margin across different model scales. Its superiority is especially significant when the model is extremely tiny, which demonstrates the effectiveness of \ourop~token mixer on information fusion at very low costs.
\ours, \ours-T, and \ours-ET models achieve 4202, 5304, and 7470 images/s thoughtput on ImageNet-1K tested with one NVIDIA A100 GPU, respectively, which is 13.5\%, 8.2\%, and 14.9\% faster than MobileViT-S/XS/XXS. More detailed results are in the Supplementary.

\begin{table}[htbp]
  \centering
  \resizebox{\linewidth}{!}{
    \begin{tabular}{l|cc|ccc}
    \toprule
    \multirow{3}[0]{*}{\quad\quad Model} & \multicolumn{2}{c|}{Detection} & \multicolumn{3}{c}{Segmentation} \\
    & \multirow{2}[0]{*}{Param.}\!& mAP(\%) & \multirow{2}[0]{*}{Param.} & \multicolumn{2}{c}{mIOU(\%)} \\
    &      & COCO   &      & ADE20K & VOC \\
    \midrule
    MViT-XXS~\cite{mehta2021mobilevit} & 1.9   & 18.5  & 1.9   &    -   & 73.6 \\
    MViTv2-0.5~\cite{mobilevitv2} & 2.0   & 21.2  & 3.6   & 31.2  & 75.1 \\
    \rowcolor[gray]{0.9}\ours-ET & 1.9   & 21.8  &  2.2    & 33.0    & 76.1 \\
    \midrule
    MViT-XS~\cite{mehta2021mobilevit} & 2.7   & 24.8  & 2.9   &   -    & 77.1 \\
    MViTv2-0.75~\cite{mobilevitv2} & 3.6   & 24.6  & 6.2   & 34.7  & 75.1 \\
    \rowcolor[gray]{0.9}\ours-T & 3.0   & 25.3  &  3.5    & 36.9  & 77.8 \\
    \midrule
    ResNet-50~\cite{he2016deep} & 22.9  & 25.2  & 68.2  & 36.2  & 76.8 \\
    \hline
    MNetv1~\cite{howard2017mobilenets} & 5.1   & 22.2  & 11.2  &  -     & 75.3 \\
    MNetv2~\cite{sandler2018mobilenetv2} & 4.3   & 22.1  & 18.7   & 34.1  & 75.7 \\
    MViT-S~\cite{mehta2021mobilevit} & 5.7   & 27.7  & 6.4   &   -    & 79.1 \\
    MViTv2-1.0~\cite{mobilevitv2} & 5.6   & 26.5  & 9.4  & 37.0  & 78.9 \\
    EdgeNext~\cite{maaz2023edgenext} & 6.2   & 27.9  & 6.5   &  -     & 80.2 \\
    \rowcolor[gray]{0.9}\ours~ & 5.6   & 28.4  & 6.9   & 38.4  & 80.5 \\
    \bottomrule
    \end{tabular}%
    }
    \vspace{+1.6mm}
  \caption{Comparisons of our \ours~with other state-of-the-art models for object detection on COCO dataset, and segmentation on ADE20k and VOC dataset. Here, Param., MNet and MViT are short for Paramters, MobileNet and MobileViT, respectively.}
  \label{tab:sota_downstream}%
\end{table}%

\subsection{Object Detection}

\paragraph{Settings.}
We conduct object detection experiments on MS-COCO dataset~\cite{lin2014microsoft}, Following the common practices in~\cite{howard2017mobilenets,sandler2018mobilenetv2,mehta2021mobilevit,mobilevitv2, maaz2023edgenext}, we compare different lightweight backbones upon the Single Shot Detection (SSD)~\cite{liu2016ssd} framework wherein separable convolutions are adopted to replace the standard convolutions in the detection head for evaluation in the lightweight setting. In the training, we load ImageNet-1K pre-trained weights as the initialization of the backbone network, and fine-tune the entire model on the training set of MS-COCO with the AdamW optimizer for 200 epochs. The input resolution of the images is 320$\times$320. Detailed introduction for the used dataset and more implementation details are in the Supplementary.

\vspace{-4mm}
\paragraph{Results.} As shown in Table \ref{tab:sota_downstream}, the detection models equipped with \ours~consistently outperforms other lightweight CNNs or transformers based detectors in mAP across different model scales. Specifically, \ours~surpasses the second-best EdgeNext~\cite{maaz2023edgenext} by 0.5\% in mAP with 0.6M fewer parameters, and surpasses the model with ResNet-50 backbone by 3.2\% in mAP using about 1/4 of parameters. Our smallest model \ours-ET outperforms the second-best model with comparable parameters MobileViTv2-0.5~\cite{mobilevitv2} by 0.6\% in mAP with fewer parameters. These results demonstrate the effectiveness of our proposed method on capturing spatial location information required by the task of object detection at low costs.

\subsection{Semantic Segmentation}

\paragraph{Settings.}
We conduct semantic segmentation experiments on two benchmarks datasets ADE20k~\cite{zhou2019semantic} and PASCAL VOC 2012~\cite{everingham2015pascal} (abbreviated as VOC). For the experiments on VOC dataset, we follow the common practices in \cite{chen2017rethinking,mehta2019espnetv2} to extend the training data with 
more annotations and data from~\cite{hariharan2011semantic} and \cite{lin2014microsoft}, respectively. The widely used semantic segmentation framework DeepLabv3~\cite{chen2017rethinking} is employed for experiments with different backbones. The input resolution of the images is set to 512$\times$512 and the ImageNet-1K pretrained weights are loaded as model initialization. All models were trained for 120 and 50 epochs on ADE20K and VOC dataset, respectively. Please see our Supplementary for more detailed introduction.

\vspace{-4mm}
\paragraph{Results.}
As the results shown in Table ~\ref{tab:sota_downstream}, \ours~performs clearly better than other lightweight networks on these two datasets. Our \ours~outperforms the second-best lightweight network MobileViTv2-1.0~\cite{mobilevitv2} by 1.4\% in mIOU on ADE20K, and outperforms the second-best lightweight model EdgeNext~\cite{maaz2023edgenext} by 0.3\% in mIOU on VOC. Besides, it achieves large improvements (2.2\% mIOU on ADE20K, 3.7\% mIOU on VOC) relative to the representative CNN model (\ieno, ResNet-50) with about 10\% of the parameters of ResNet-50. These exhibit the effectiveness of our proposed method on dense prediction tasks.

\begin{table}[t]
  \centering
  \resizebox{\linewidth}{!}{
    \begin{tabular}{l|cc|c}
    \toprule
    \multicolumn{1}{c|}{Method} & Param (M) & FLOPs (G) & Top-1 \\
    \midrule
    Base. & 5.2  & 1.3   & 77.9 \\
    Base. + Conv-mixer (3$\times$3) & 10.7   & 2.7   & 78.6 \\
    Base. + AFF w/o FFT  & 5.5   & 1.5   & 78.4 \\
    \rowcolor[gray]{0.9}Base. + \ourop~ (Our \ours)& 5.5   & 1.5   & 79.8 \\
    \bottomrule
    \end{tabular}%
    } 
    \vspace{+1.6mm}
  \caption{Comparisons of our proposed model with baseline (no spatial token mixer) and models with other token mixers in the original domain on ImageNet-1K classification. ``Base.'' denotes the baseline model discarding all \ourop~token mixers. ``Conv-Mixer (3$\times$3)'' refers to adopting token mixers implemented by 3$\times$3 convolutions in the original space. ``AFF w/o FFT'' denotes performing adaptive filtering in the original space with the same networks by discarding the Fourier transforms where ``w/o'' and ``AFF'' are short for ``without'' and ``\ourop~token mixer'', respectively.}
  \label{tab:aff2baseline}
\end{table}%

\subsection{Ablation Study}

\paragraph{Effectiveness and complexity of \ourop~token mixer.}
We analyze the effectiveness and complexity of our proposed \ourop~token mixer by comparing \ours~with the \textit{Base.} model in which all \ourop~token mixers are replaced with identity functions. As shown in Table \ref{tab:aff2baseline}, all \ourop~token mixers in \ours~only requires 0.3M parameter increase ($<6\%$) and 0.2G FLOPs increase ($\sim15\%$) relative to the baseline and improves the Top-1 accuracy on ImageNet-1K by 1.9\%. Comparing to the model with one 3$\times$3 convolution layer as the token mixer, \ieno, \textit{Base.+Conv-Mixer (3$\times$3)}, \ours~delivers 1.2\% Top-1 accuracy improvements with about only half of parameters and FLOPs. This strongly demonstrates the effectiveness and efficiency of our proposed method for token mixing in lightweight networks.


\vspace{-2mm}
\paragraph{Original vs. frequency domain.}
We compare applying the same adaptive filtering operations in original domain and in frequency domain. We discard the all Fourier and inverse Fourier transforms and remain others the same as \ours, \ieno, \textit{Base.+AFF w/o FFT} in Table \ref{tab:aff2baseline}. Our \ours~clearly outperforms it by 1.4\% Top-1 accuracy with the same model complexity. Applying adaptive filtering in the original domain is even weaker than convolutional token mixer, which indicates that only adaptive \textit{\textbf{frequency}} filters can serve as effeicient global token mixers.

\begin{table}[t]
  \centering
  \resizebox{\linewidth}{!}{
    \begin{tabular}{l|cc|c}
    \toprule
    \multicolumn{1}{c|}{Method} & Param (M) & FLOPs (G) & Top-1 \\
    \midrule
    Base. & 5.2  & 1.3   & 77.9 \\
    Base. + AFNO~\cite{guibas2021adaptive} & 5.5 & 1.5   & 78.8 \\
    Base. + GFN~\cite{rao2021global} & 6.5 & 1.5   & 79.1 \\
    Base. + FFC~\cite{chi2020fast} & 7.7 & 1.7   & 79.1 \\
    Base. + DFF~\cite{lin2022deep} & 7.7 & 1.7   & 79.3 \\
    Base. + FNO~\cite{li2020fourier} & 141.0 & 1.5   & 79.7 \\
    Base. + AFF w. SUM & 5.5 & 1.5  & 78.8 \\
    \rowcolor[gray]{0.9}Base. + \ourop~(\ours) & 5.5 & 1.5   & 79.8 \\
    \bottomrule
    \end{tabular}%
    }
    \vspace{+1mm}
  \caption{Comparisons of our design for \ourop~token mixer and other frequency-domain operations in previous works \cite{li2020fourier, rao2021global, chi2020fast, lin2022deep, guibas2021adaptive} in terms of their roles for token mixing on ImageNet-1K. ``AFF w. SUM'' denotes replacing the Hadamard product with a summation operation, ``w.'' is short for ``with''. }
  \label{tab:freq_operations}
\end{table}%

\vspace{-2mm}
\paragraph{Comparisons of different frequency operations.} 
We compare the frequency operation design in \ourop~token mixer with those in previous works~\cite{li2020fourier,rao2021global,chi2020fast,lin2022deep,guibas2021adaptive} in terms of their effects as token mixers. The results are in Table~\ref{tab:freq_operations}. As analyzed in Sec.\ref{sec:analysis}, FFC~\cite{chi2020fast} and AFNO~\cite{guibas2021adaptive} actually perform filtering with deterministic functions, resulting in the lack of the adaptivity to semantics. The frequency-domain operations in them are both obviously inferior to ours. Moreover, our operation design is also clearly better than those in GFN~\cite{rao2021global} and FNO~\cite{li2020fourier} since they perform filtering with network parameters implemented masks. These masks are fixed after training and lead to a large increase in parameters (\textit{Base.+FNO} has more than 25$\times$ parameters as ours). Note that the implementation of FNO~\cite{li2020fourier} with unshared fully connected layers for each frequency component results in a significant increase in the number of parameters. DFF~\cite{lin2022deep} is designed for filtering out the frequency components adverse to domain generalization, thus requiring a spatial mask only. Our \ours~is superior to \textit{Base.+DFF} by 0.5\% with fewer parameters and FLOPs, demonstrating the importance of channel-wise mixing. This will be further verified with a fairer comparison. These existing frequency-domain operations might be similar with our proposed one at the first glance, but they are designed for different purposes and perform worse than ours as token mixers. When replacing the Hadamard product in our method with a summation operation, the Top-1 accuracy drops by 1.0\% since the equivalence introduced in Sec.\ref{sec:AFF-token-mixer} no longer holds.

\vspace{-2mm}
\paragraph{The importance of channel-specific token mixing.}
We have preliminarily demonstrated this by comparing the frequency-domain operations in DFF \cite{lin2022deep} and ours. Considering their masks are learned with different networks, here, we conduct a fairer comparison by adopting an average pooling along the channels of the learned masks in \ourop~token mixer. As shown in Table \ref{tab:channel-wise}, frequency filtering with the masks of a shape of 1$\times$H$\times$W lead to 0.5\% accuracy drop with the same model complexities, verifying the importance of channel-specific token mixing. This is because different semantic attributes of the learned latent representations distribute in different channels \cite{bau2020understanding,wu2021stylespace}, thus requiring channel-specific weights for token mixing. 
Besides, it delivers the same accuracy with \textit{Base.+DFF} in Table~\ref{tab:freq_operations}. This indicates that the network architectures here are in fact not dominating factors for the effects of token mixing, allowing us to use a lightweight one.

\begin{table}[t]
  \centering
    \resizebox{0.8\linewidth}{!}{
    \begin{tabular}{l|cc|c}
    \toprule
    \multicolumn{1}{c|}{Mask Shape}  & Param (M) & FLOPs (G) & Top-1 \\
    \midrule
     1$\times$H$\times$W & 5.5   & 1.5   & 79.3 \\
     \rowcolor[gray]{0.9}C$\times$H$\times$W & 5.5   & 1.5   & 79.8 \\
    \bottomrule
    \end{tabular}%
    }
    \vspace{+1mm}
    \caption{Experiments of verifying the importance of channel-specific token mixing on ImageNet-1K. Here, we adopt an average pooling operation along the channel dimension of the masks learned in \ours, yielding the mask with a shape of 1$\times$H$\times$W. This mask is shared across channels.}
  \label{tab:channel-wise}%
\end{table}%

\vspace{-2mm}
\paragraph{Comparisons of hyper-parameter choices.}
As shown in Fig.\ref{fig:framework}, we adopt two group linear layers (also known as 1$\times$1 convolution layers) with ReLU to learn the masks for our proposed adaptive frequency filtering. As shown in Table \ref{tab:mask_gen}, improving the kernel size cannot further improve the performance but leads to larger model complexities. Moreover, we keep the spatial kernel size as 1$\times$1 while using different group numbers. When $N_{group}$=$C$, the Top-1 accuracy drops by 0.4\%, in which depthwise convolutions are used so that the contexts among different channels are under-exploited for inferring the weights of token mixing. When $N_{group}$=1, it means that regular convolution/linear layers are used, which slightly improve the Top-1 accuracy by 0.1\% at the expense of 40\% parameters increase and 33.3\% FLOPs increase. This setting explores more contexts but results in a worse accuracy and efficiency trade-off.

\begin{table}[t]
  \centering
  \resizebox{0.9\linewidth}{!}{
    \begin{tabular}{c|c|cc|c}
    \toprule
    \multicolumn{1}{c|}{Spatial K-Size} & $N_{group}$ & Param. (M) & FLOPs (G) & Top-1 \\
    \midrule
    $1 \times 1$ & $C$        & 5.3   & 1.4   & 79.4 \\
    $1 \times 1$ & 1        & 7.7   & 2.0   & 79.9 \\
    $3 \times 3$ & 8  &  7.9     &  2.0     & 79.8 \\
    \rowcolor[gray]{0.9}$1 \times 1$ & 8        & 5.5   & 1.5   & 79.8 \\
    \bottomrule
    \end{tabular}%
    }
    \vspace{+1mm}
  \caption{Comparisons of different hyper-parameter choices in the sub-network for learning the filtering masks in \ours~on ImageNet-1K. ``Spatial K-Size'' refers to the spatial size of convolution kernels. $N_{group}$ denotes the number of groups for group linear or convolution layers. $C$ is the total number of channels.}
  \label{tab:mask_gen}%
\end{table}%


\section{Conclusion}
In this work, we reveal that \textit{adaptive frequency filters can serve as efficient global token mixers} in a mathematically equivalent manner. Upon this, we propose Adaptive Frequency Filtering (\ourop) token mixer to achieve low-cost adaptive token mixing in the global scope. Moreover, we take \ourop~token mixers as primary neural operators to build a lightweight backbone network, dubbed \ours. \ours~achieves SOTA accuracy and efficiency trade-offs compared to other lightweight network designs across multiple vision tasks. Besides, we revisit the existing frequency-domain neural operations for figuring out what matters in their designs for token mixing. We hope this work could inspire more interplay between conventional signal processing and deep learning technologies.

{\small
\bibliographystyle{ieee_fullname}
\bibliography{egbib}
}

\clearpage
\section*{\Large{\textbf{Supplementary Material}}}

\section{Detailed Network Architectures}

As introduced in our manuscript, we build three versions of our proposed hierarchical backbone \ours~with different channel dimensions, namely \ours, \ours-T and \ours-ET, respectively. Here, we provide the detailed model configurations of them in Table~\ref{tab:model_detail}. Specifically, following commonly used designs~\cite{mehta2021mobilevit, mobilevitv2}, we adopt a convolution stem for tokenization, which consists of a 3$\times$3 convolution layer with a stride of 2, followed by four MBConv layers. MBConv is short for the Mobile Convolution Block in ~\cite{sandler2018mobilenetv2} with a kernel size of 3. After tokenization, three stages are cascaded as the main body of \ours, where each stage is composed of a MBConv layer with stride 2 for down-sampling in spatial and $N_i$ \ourblock. Specifically, we set $N_1=2$, $N_2=4$ and $N_3=3$.

\begin{table*}[t]
  \centering
    \begin{tabular}{l|c|c|c|ccc}
    \toprule
    \multicolumn{1}{c|}{\multirow{2}[0]{*}{Layer / Block}} & \multirow{2}[0]{*}{Resolution} & \multirow{2}[0]{*}{Down-sample Ratio} & \multirow{2}[0]{*}{Number of Blocks} & \multicolumn{3}{c}{Number of Channels} \\
          &       &       &       & \ours-ET   & \ours-T    & \ours \\
    \midrule
    Image & $256^2$ &   -    & 1     & 16    & 16    & 16 \\
    \midrule
    \multirow{2}[0]{*}{Conv Stem} & $128^2$ & $\downarrow$ 2     & 1     & 32    & 32    & 32 \\
     & $64^2$ & $\downarrow$ 2     & 4     & 48    & 48    & 64 \\
    \midrule
    Down-sampling & $32^2$ & $\downarrow$ 2     & 1     & 64    & 96    & 128 \\
    \ourblock & $32^2$ & -     & 2     & 64    & 96    & 128 \\
    \midrule
    Down-sampling & $16^2$ & $\downarrow$ 2     & 1     & 104   & 160   & 256 \\
    \ourblock & $16^2$ & -     & 4     & 104   & 160   & 256 \\
    \midrule
    Down-sampling & $8^2$   & $\downarrow$ 2     & 1     & 144   & 192   & 320 \\
    \ourblock & $8^2$   & -     & 3     & 144   & 192   & 320 \\
    \midrule
    Parameters &   -    &   -    &  -     & 1.4M   & 2.6M   & 5.5M \\
    FLOPs &  -     &   -    &   -    & 0.4G   & 0.8G   & 1.5G \\
    \bottomrule
    \end{tabular}%
  \vspace{+1.6mm}
  \caption{Detailed model configurations. The resolution and the number of channels in above table correspond to the output representations for each layer/block.}
  \label{tab:model_detail}%
\end{table*}%

\section{Detailed Introduction for Dataset}

\textbf{ImageNet~\cite{russakovsky2015imagenet}} is a large-scale dataset with over 1.2 million images and 1000 object categories for the visual recognition challenge. It serves as the most widely used dataset for image classification. The images in this dataset are of varying sizes and resolutions, and include various objects in diverse backgrounds. We train our models on Imagenet-1k dataset from scratch to illustrate the effectiveness and efficiency of our proposed models on image classification.


\textbf{MS-COCO~\cite{lin2014microsoft}} (abbreviated as COCO) is a widely used benchmark dataset for object detection, instance segmentation, and keypoint detection tasks. It contains more than 200,000 images and 80 object categories, annotated with bounding boxes, masks, and keypoints. The objects in this dataset are diverse and challenging, including people, animals, vehicles, household items, \etcno.


\textbf{ADE20k~\cite{zhou2019semantic}} is a dataset consisting of 20,210 images covering a wide range of indoor and outdoor scenes. The images in this dataset are annotated with pixel-level labels for 150 semantic categories, such as sky, road, person and so on. This dataset is widely used for evaluating the performance of deep models on semantic segmentation and scene understanding.


\textbf{PASCAL VOC 2012~\cite{everingham2015pascal}} (abbreviated as VOC) is a widely used benchmark for object recognition, object detection, and semantic segmentation. It consists of 20 object categories and contains more than 11,000 images with pixel-level annotations for object boundaries and semantic categories. This dataset is challenging due to the large variability in object appearances and the presence of occlusions and clutter within it. 


\section{Detailed Experiment Settings}

We provide detailed experiment settings for different tasks in Table~\ref{tab:train_config}, including the detailed configurations for model, data and training.
\begin{table*}[t]
  \centering
  \resizebox{\linewidth}{!}{
    \begin{tabular}{l|ccc|c|cc}
    \toprule
    \multicolumn{1}{c}{Task}  & \multicolumn{3}{|c}{Image Classification} & \multicolumn{1}{|c}{Object Detection} & \multicolumn{2}{|c}{Semantic Segmentation} \\
    \midrule
    Model & \ours-ET & \ours-T & \ours   & \ours   & \ours   & \ours \\
        \midrule
    EMA & \ding{51}     & \ding{51}     & \ding{51}     & \ding{51}     & \ding{51}     & \ding{51} \\
    Weight Initialization & \makecell[c]{Kaiming \\ normal} & \makecell[c]{Kaiming \\ normal} & \makecell[c]{Kaiming \\ normal} & \makecell[c]{ImageNet-1k\\ pretrain} & \makecell[c]{ImageNet-1k\\ pretrain} & \makecell[c]{ImageNet-1k\\ pretrain} \\
    \midrule
    Dataset & ImageNet-1k & ImageNet-1k & ImageNet-1k & COCO  & ADE20k & PASCAL VOC \\
    Resolution & $256^2$ & $256^2$ & $256^2$ & $320^2$ & $512^2$ & $512^2$ \\
    \midrule
    RandAug & \ding{55}     & \ding{55}     & \ding{51}     & \ding{55}     & \ding{55}     & \ding{55} \\
    CutMix & \ding{55}     & \ding{55}     & \ding{51}     & \ding{55}     & \ding{55}     & \ding{55} \\
    MixUp & \ding{55}     & \ding{55}     & \ding{51}     & \ding{55}     & \ding{55}     & \ding{55} \\
    Random Resized Crop & \ding{51}     & \ding{51}     & \ding{51}     & \ding{55}     & \ding{51}     & \ding{51} \\
    Random Horizontal Flip & \ding{51}     & \ding{51}     & \ding{51}     & \ding{55}     & \ding{51}     & \ding{51} \\
    Random Erase & \ding{55}     & \ding{55}     & \ding{51}     & \ding{55}     & \ding{55}     & \ding{55} \\
    Gaussian Noise & \ding{55}     & \ding{55}     & \ding{55}     & \ding{55}     & \ding{51}     & \ding{51} \\
    \midrule
    Label Smoothing & \ding{51}     & \ding{51}     & \ding{51}     & \ding{55}     & \ding{55}     & \ding{55} \\
    Loss  & CE    & CE    & CE    & Ssd Multibox    & CE    & CE \\
    Optimizer & AdamW & AdamW & AdamW & AdamW & AdamW & AdamW \\
    Weight Decay & 0.008 & 0.02  & 0.05  & 0.05  & 0.05  & 0.05 \\
    Warm-up Iterations & 20 k  & 20 K  & 20 k  & 500   & 500   & 500 \\
    LR Scheduler & Cosine & Cosine & Cosine & Cosine & Cosine & Cosine \\
    Base LR & 0.009 & 0.0049 & 0.002 & 0.0007 & 0.0005 & 0.0005 \\
    Minimal LR & 0.0009 & 0.00049 & 0.0002 & 0.00007 & 1.00E-06 & 1.00E-06 \\
    Number of Epochs & 300   & 300   & 300   & 200   & 120   & 50 \\
    Batch Size & 1024  & 1024  & 1024  & 128   & 16    & 128 \\
    \bottomrule
    \end{tabular}%
    }
    \vspace{+0.6mm}
    \caption{Detailed training configurations of \ours, \ours-T, and \ours-ET models on different tasks. ``LR'' denotes the learning rate and ``EMA'' is short for Exponential Moving Average. For object detection and semantic segmentation tasks, \ours-T and \ours-ET use the same training configuration as \ours.}
  \label{tab:train_config}%
\end{table*}%

\section{More Experiment Results}

\subsection{Quantitative Results}

\paragraph{Running speed evaluation.}
We report the model speeds of our proposed \ours~models on mobile devices (iPhone) and GPUs, and compare them with other advanced lightweight models that incorporate global token mixers in Table~\ref{tab:lantency}. Models with similar Top-1 accuracy are grouped together for clear comparison. The latency results are equivalently measured by CoreML\protect\footnote{\url{https://github.com/apple/coremltools}} on an iPhone with a batch size of 1. The throughput results are measured with TorchScript\protect\footnote{\url{https://github.com/pytorch/pytorch/blob/master/torch/csrc/jit/OVERVIEW.md}} on an A100 GPU (batch size = 128). As shown in Table~\ref{tab:lantency}, thanks to the \ourop~token mixer, \ours~outperforms other network designs by a clear margin across different model scales. On GPUs (NVIDIA A100), \ours~achieves 0.4\% Top-1 accuracy improvement with 179 image/s lager throughput compared to the second fastest model EdgeNext-S. On the mobile device (iPhone), \ours~also surpasses the second fastest model mobilevitv2 by 1.7\% Top-1 accuracy with 0.3 ms less latency. These results reflect high effectiveness and efficiency of our proposed method.

\begin{table}[t]
  \centering
    \resizebox{\linewidth}{!}{
    \begin{tabular}{l|cc|cc|c}
    \toprule
     \multicolumn{1}{c|}{Model} & \makecell[c]{Param.  \\ (M)} & \makecell[c]{FLOPs  \\ (G)} & \makecell[c]{Latency  \\ (ms)} & \makecell[c]{Throughput  \\ (images/s)} & Top-1 \\
    \midrule
    MViT-XXS~\cite{mehta2021mobilevit} & 1.3 & 0.4   & 4.8   & 6803  & 69.0 \\
    MViTv2-0.5~\cite{mobilevitv2} & 1.4 & 0.5   & 1.6   & 7021  & 70.2 \\
    EdgeNext-XXS~\cite{maaz2023edgenext} & 1.3 & 0.3 & 1.7 & 7768 & 71.2 \\
    \rowcolor[gray]{0.9}\ours-ET & 1.4 & 0.4  &  1.4     & 8196  & 73.0 \\
    \midrule
    MViT-XS~\cite{mehta2021mobilevit} & 2.3 & 1.0 &  7.0   & 4966  & 74.8 \\
    MViTv2-0.75~\cite{mobilevitv2} & 2.9  & 1.0 &   2.4    & 5150  & 75.6 \\
    EdgeNext-XS~\cite{maaz2023edgenext} & 2.3 & 0.5 & 2.6 & 5307 & 75.0 \\
    \rowcolor[gray]{0.9}\ours-T & 2.6 & 0.8   &  2.1     & 5412  & 77.0 \\
    \midrule
    CycleMLP-B1~\cite{chen2021cyclemlp} & 15.2 & 2.1 & 15.2 & 3073 & 79.1 \\
    PoolFormer-S12~\cite{yu2022metaformer} & 11.9 & 1.8 & 5.3 & 3922 & 77.2 \\
    MFormer-294~\cite{chen2022mobile}  & 11.8 & 0.3  & 40.7  & 2790  & 77.9 \\
    MViT-S~\cite{mehta2021mobilevit} & 5.6 & 2.0  & 9.9   & 3703  & 78.4 \\
    MViTv2-1.0~\cite{mobilevitv2} & 4.9 & 1.8   & 3.4   & 3973  & 78.1 \\
    EdgeNext-S~\cite{maaz2023edgenext}  & 5.6 & 1.3 & 6.4 & 4023 & 79.4 \\
    \rowcolor[gray]{0.9}\ours & 5.5 & 1.5   &  3.1    & 4202  & 79.8 \\
    \bottomrule
    \end{tabular}}%
    \vspace{+1.6mm}
  \caption{Results of model speed evaluation. Here, the latency results are equivalently measured using CoreML on an iPhone with a batch size of 1. The throughput results are measured using TorchScript on an A100 GPU with a batch size of 128.}
  \label{tab:lantency}%
\end{table}%

\paragraph{Evaluation on more downstream task frameworks.}
For the experiments reported in our main paper (\egno, Table~\ref{tab:sota_downstream}), we adopt the most commonly used task frameworks, \ieno, SSD and Deeplabv3, in accordance with recent studies~\cite{sandler2018mobilenetv2,mehta2021mobilevit,mobilevitv2,maaz2023edgenext} on general-purpose lightweight backbone design to ensure a fair comparison. Moreover, to evaluate the compatibility of \ours~with more downstream task frameworks, we incorporated AFFNet into more downstream task frameworks ~\cite{ge2021yolox,tan2020efficientdet,guo2022visual,yang2022moat} as their encoders. These frameworks involve multi-stage/scale feature interactions via some task-specific architecture designs. By utilizing AFFNet as the encoders, these models perform consistently better compared to their vanilla versions in mAP@COCO and mIOU@ADE20K, as presented in Table~\ref{tab:downstream2}. There results further demonstrate that our proposed \ours~is compatible with diverse downstream task frameworks and generally applicable.

\begin{table}[!h]
\small
\centering

\resizebox{\linewidth}{!}{
\begin{tabular}{c|cc|cc}
\toprule
Task    & \multicolumn{2}{c|}{Detection(mAP)} & \multicolumn{2}{c}{Segmentation(mIOU)}  \\
Framework From &  yolox~\cite{ge2021yolox} & efficientdet~\cite{tan2020efficientdet}  & van~\cite{guo2022visual} & moat~\cite{yang2022moat} \\ 
\midrule
w. Origin Encoder  & 32.8  & 40.2  & 38.5 & 41.2  \\
\rowcolor[gray]{0.9} w. \ours~Encoder & 35.9  & 41.6  & 43.2 & 41.5  \\
\bottomrule
\end{tabular}
}
\vspace{+0.8mm}
\caption{Performance evaluation on more downstream task frameworks. Our proposed \ours~are integrated into them as their encoders to compare with their original ones.}

\label{tab:downstream2}
\end{table}

\paragraph{Comparisons of different frequency transforms.} 
We investigate the effectiveness of adopting different frequency transforms in implementing our proposed \ourop~token mixer. Specifically, we compare using FFT and using wavelet transform or Discrete Cosine Transform (DCT). The comparison results are in Table~\ref{tab:fourier_process}. We observe that adopting the wavelet transform also attains improvements compared to the baseline model without any frequency transforms, but it is clearly inferior to adopting FFT as we recommend. This is because the wavelet transform is a low-frequency transformation that performs our proposed filtering operation in a local space, which limits the benefits of our \ourop~token mixer as a global token mixer. Moreover, DCT is slightly inferior to FFT since that DCT is a Fourier-related transform with coarser transform basis. It thus leads to more information loss when mixing tokens. Besides, DCT only performs transformation only on real numbers.

\begin{table}[t]
  \centering
  \resizebox{0.8\linewidth}{!}{
    \begin{tabular}{l|cc|c}
    \toprule
    \multicolumn{1}{c|}{\makecell[c]{Frequency  \\ Transformations}} & Param (M) & FLOPs (G) & Top-1 \\
    \midrule
    Baseline &   5.5 & 1.5    &  78.4 \\
    Wavelet &   5.5 & 1.5    &  78.6 \\
    DCT     &   5.5 & 1.5    &  79.6 \\
    \rowcolor[gray]{0.9}FFT (Ours)  &  5.5 & 1.5    &  79.8 \\
    \bottomrule
    \end{tabular}%
    }
  \vspace{+2.3mm}
  \caption{Comparisons of adopting different frequency transforms in implementating our proposed method. ``Baseline'' denotes the model without any frequency transforms, ``Wavelet'' denotes the wavelet transforms with the Haar filters, and ``DCT'' is short for Discrete Cosine transform.}
  \label{tab:fourier_process}%
\end{table}%

\paragraph{The order of token-mixing and channel-mixing.}
We study the effect of the order of token mixing and channel mixing in backbone design. As shown in Table~\ref{tab:aff_block}, \textit{channel-mixing first} design is slightly superior to the \textit{token-mixing first} design, indicating it would be better to perform within-token refinement before token mixing. Overall, they deliver very close results.

\begin{table}[t]
  \centering
  \resizebox{0.9\linewidth}{!}{
    \begin{tabular}{l|cc|c}
    \toprule
    \multicolumn{1}{c|}{Order} & Param (M) & FLOPs (G) & Top-1 \\
    \midrule
    Token-mixing first &   5.5    & 1.5 &  79.7 \\
    \rowcolor[gray]{0.9}Channel-mixing first (Ours) &   5.5    & 1.5 &  79.8 \\
    \bottomrule
    \end{tabular}%
    }
  \vspace{+1.6mm}
  \caption{Investigation results of the effects of the order of token-mixing and channel-mixing in \ourblock. ``Token-mixing first'' denotes performing token mixing before channel mixing while ``Channel-mixing first'' is an opposite order.}
  \label{tab:aff_block}%
\end{table}%

\paragraph{The design of channel mixer.}
In this paper, we focus on the design of token mixer while the channel mixer is not the main point of this work. Thus, we employ a plain channel mixer implemented by Mobilenet Convolution Block (MBConv) \cite{sandler2018mobilenetv2} following prior works~\cite{wang2022pvt,chen2021regionvit,tu2022maxvit,yang2022moat}. Here, we compare two dominated designs of the channel mixer in Table~\ref{tab:channel_mixer} for a detailed empirical study. 
Feed-Forward Network (FFN)~\cite{vaswani2017attention,dosovitskiy2020image} adopts two cascaded linear layers while MBConv adds a depth-wise 3$\times$3 convolution layer between two linear layers. We find MBConv is more powerful as the channel mixer in lightweight neural network design than FFN, in which their computational costs are almost the same.

\begin{table}[t]
  \centering
  \resizebox{0.9\linewidth}{!}{
    \begin{tabular}{l|cc|c}
    \toprule
    \multicolumn{1}{c|}{Channel-mixing Design} & Param (M) & FLOPs (G) & Top-1 \\
    \midrule
    FFN &   5.5    & 1.5 &  79.5 \\
    \rowcolor[gray]{0.9}MBConv (Ours) &   5.5    & 1.5 &  79.8 \\
    \bottomrule
    \end{tabular}%
    }
  \vspace{+1.6mm}
  \caption{Comparisons of two mainstream designs for channel mixers. They are  FFN (Feed-Forward Network) and MBConv (Mobilenet Convolution Block) as channel mixer. Note that the design of channel mixers is not the focus of our work, and we adopt MBConv as token mixers in our proposed method.}
  \label{tab:channel_mixer}%
\end{table}%


\section{Visualization Results}
\begin{figure*}[!t]
	\centering
	\includegraphics[width=\textwidth]{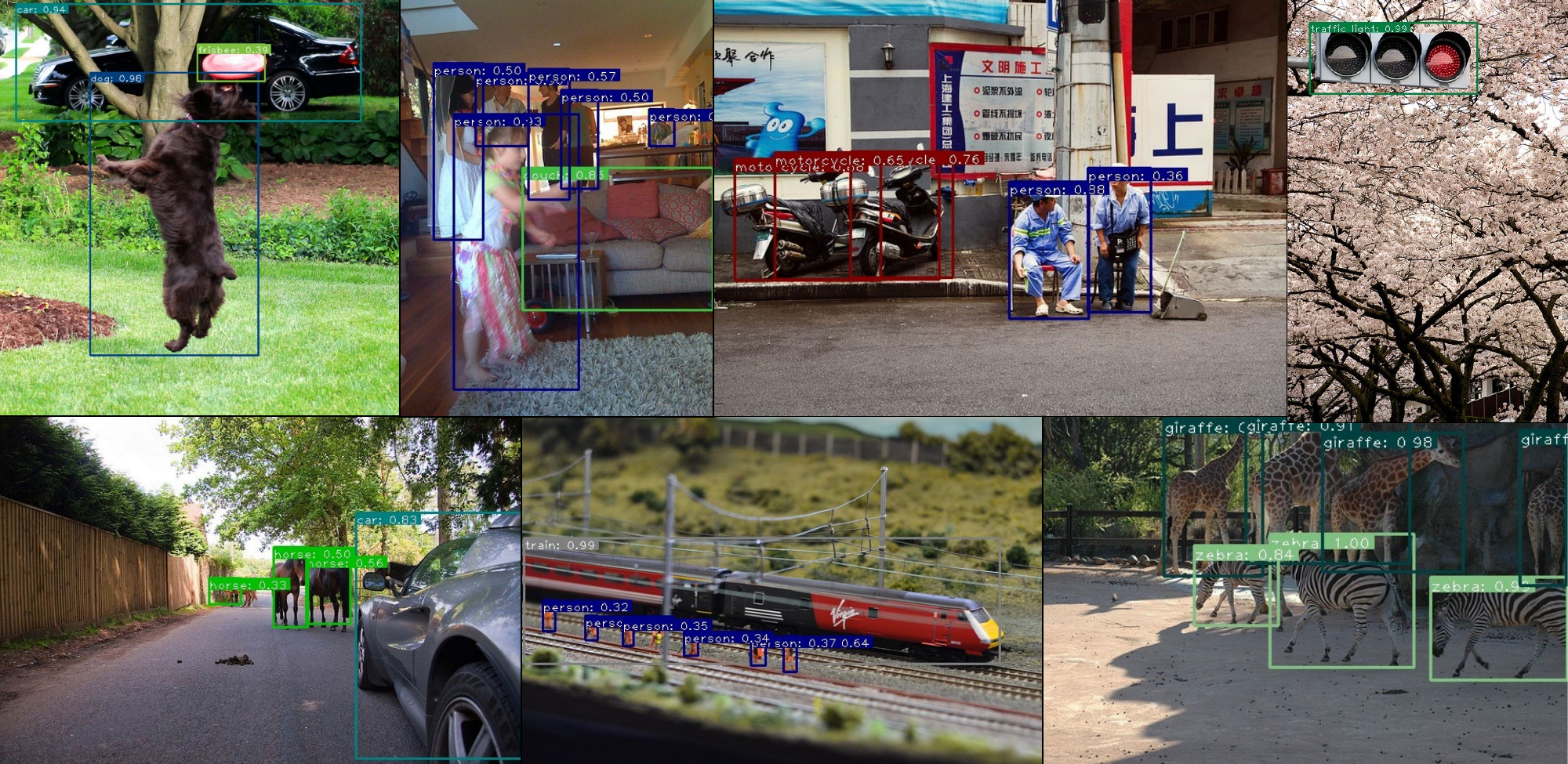}
        \caption{Qualitative results of the detection model with our \ours~as the backbone on the validation set of COCO dataset.}
	\label{fig:qualitative_results_detection}
\end{figure*}

\begin{figure*}[!t]
	\centering
	\includegraphics[width=0.7\textwidth]{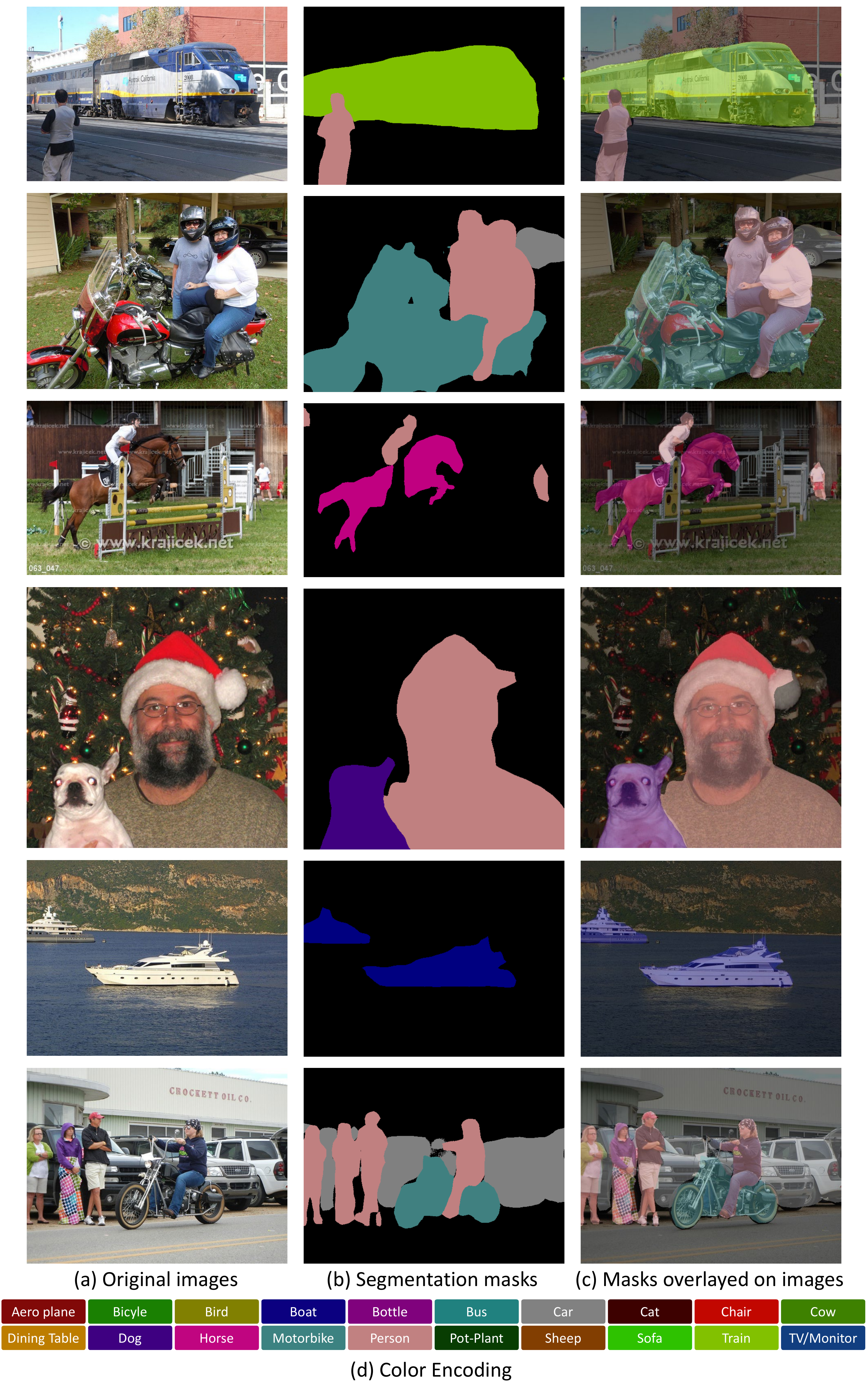}
        \caption{Qualitative results of the segmentation model with our \ours~as the backbone on unseen validation set of COCO dataset. This model is trained on the Pascal VOC dataset with 20 segmentation classes.}
	\label{fig:qualitative_results_segmentation}
\end{figure*}

We present the qualitative results of \ours~on object detection and semantic segmentation in Fig.~\ref{fig:qualitative_results_detection} and Fig.~\ref{fig:qualitative_results_segmentation}, respectively. These qualitative results demonstrate that our proposed \ours~is capable of precisely localizing and classifying objects in the dense prediction tasks with diverse object scales and complex backgrounds as a lightweight network design. And this demonstrates the effectiveness of our proposed \ourop~token mixer in preserving the spatial structure information during token mixing. 

\section{Limitations}
Although we show the superiority of \ours~in the running speed, We have to point out that there is still a gap between the current running speed and the theoretical upper limit of the speed it can achieve, as the speed optimization in engineering implementation of frequency transformations such as FFT/iFFT has not been fully considered yet. Besides, this work only focuses on the vision domain currently. We are looking forwards to its further extension to other research fields.

\end{document}